\documentclass{article}
\usepackage{ijcai17}

\usepackage{times}

\usepackage{algorithm}
\usepackage{algorithmic}
\usepackage{amsmath,bm}
\usepackage{multirow}
\usepackage{array}
\usepackage{booktabs}
\usepackage{rotating}
\usepackage{float}

\newtheorem{as}{Assumption}
\title{Zero-Shot Learning posed as a Missing Data Problem}
\author{Bo Zhao$^{1}$, Botong Wu$^{1}$, Tianfu Wu$^{2}$, Yizhou Wang$^{1}$\\
\normalsize $^1$Nat'l Engineering Laboratory for Video Technology,\\
\normalsize Key Laboratory of Machine Perception (MoE),\\
\normalsize    Cooperative Medianet Innovation Center, Shanghai,\\
\normalsize	 Sch'l of EECS, Peking University, Beijing, 100871, China\\
\normalsize	$^2$Department of ECE and the Visual Narrative Cluster, North Carolina State University \\
{\small  \{bozhao, botongwu, Yizhou.Wang\} @pku.edu.cn, tianfu\_wu@ncsu.edu}
}

\begin{document}

\maketitle

\begin{abstract}
This paper presents a method of zero-shot learning (ZSL) which poses ZSL as the missing data problem, rather than the missing label problem.  Specifically, most existing ZSL methods focus on learning mapping functions from the image feature space to the label embedding space. Whereas, the proposed method explores a simple yet effective transductive framework in the reverse way \--- our method estimates data distribution of unseen classes in the image feature space  by transferring knowledge from the label embedding space. In experiments, our method outperforms the state-of-the-art on two popular datasets.
\end{abstract}

%%%%%%%%% BODY TEXT
\section{Introduction}

\begin{figure}[h] %[htbp]
  \centering
  \includegraphics[width=0.45\textwidth]{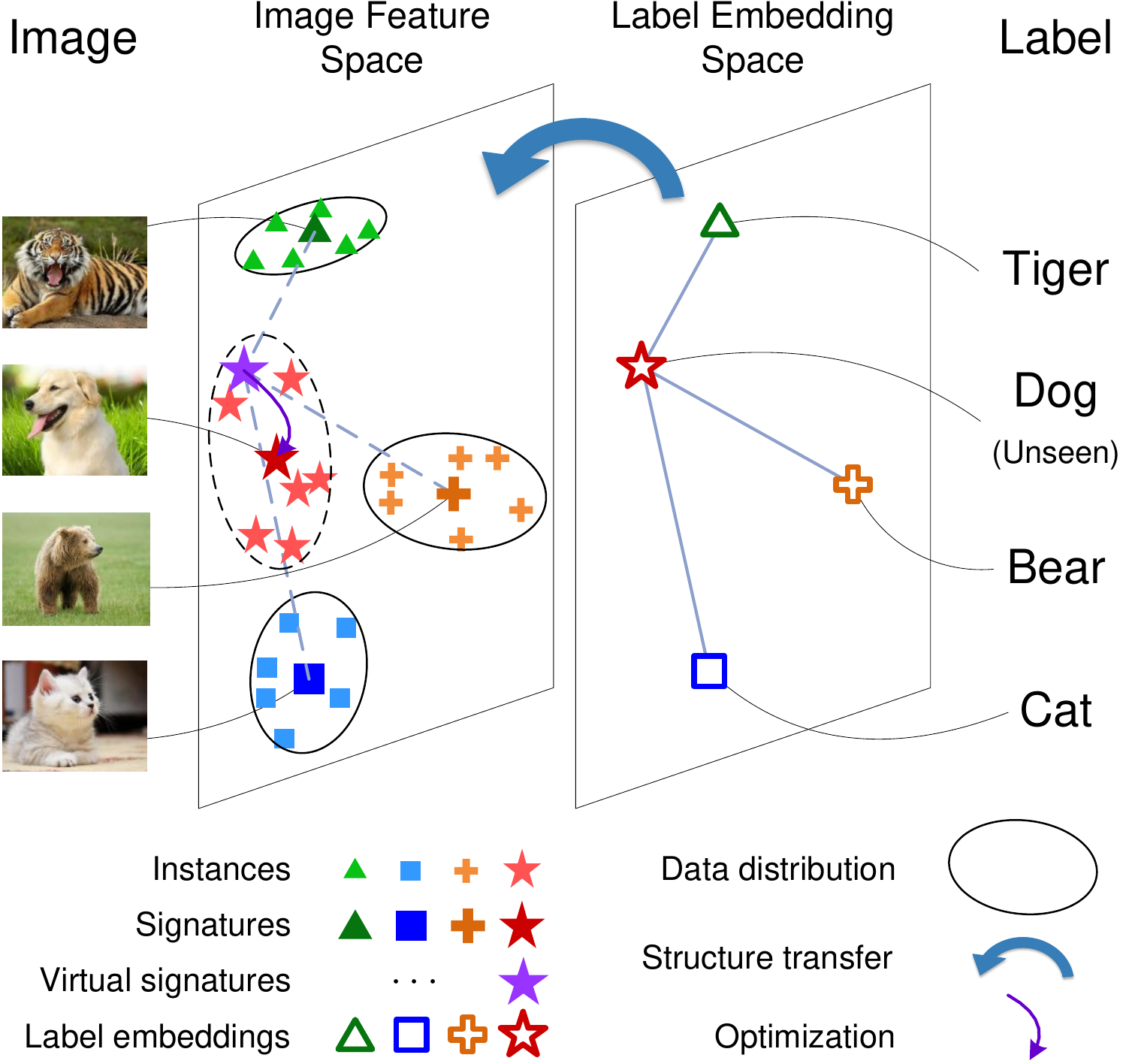}
  \caption{Illustration of the proposed method. The manifold structure (the straight lines) in the label embedding space is transferred to the image feature space for synthesizing the virtual signature (the purple star) of an unseen class. The purple arrow points to a refined signature, which demonstrates that the synthesized virtual signature is optimized after running the Expectation-Maximization algorithm so that unseen data are assigned to labels according to the data distribution.}
  \label{fig:firstfig}
\end{figure}
%In the image feature space, instances of each class are assumed to approximately subject to Gaussian distribution. All instances are jointly modeled using Gaussian mixture model. We utilize the signature to represent the data distribution of each class. In our implementation, signatures are formulated with Gaussian parameters.
%Different from popular methods, we directly classify new instances in the image feature space, rather than learn mapping function from the image feature space to the label embedding space.
% The larger red pentagram in the image feature space means the authentic signature of the unseen class (dog).

The recent success of deep learning heavily relies on a large amount of labeled training data. For some classes, e.g., rare wildlife and unusual diseases, it is expensive even impossible to collect thousands of samples. Traditional supervised learning frameworks cannot work well in this situation. Zero-shot learning (ZSL) that aims to recognize instances of an unseen class is considered to be a promising solution.

In ZSL, data are (datum, label) pairs and these data pairs are split into labeled seen classes (source domain) and unlabeled unseen classes (target domain where labels are missing). The seen classes and unseen classes are disjointed. Therefore, ``auxiliary information'' is introduced to enable knowledge transfer from seen classes to unseen ones so that given a datum from the unseen classes, its label can be predicted. Often used auxiliary information includes attributes\cite{lampert2014attribute}, textual description\cite{lei2015predicting} and word vectors of labels\cite{socher2013zero}), etc. In most practice, labels are embedded in ``label embedding space". Data (e.g., images) are embedded in (e.g., image) feature space (using hand-craft or deep learning feature extractors). In the following of this paper, we introduce ZSL in the context of image recognition.

%constructing image feature space and label embedding space according to given data pairs

One popular type of ZSL is implemented in an inductive way, i.e. models are trained on seen classes then applied to unseen classes directly. Usually, inductive ZSL includes three steps: i) embedding images and labels in the image feature space and label embedding space respectively; ii) learning the mapping function from the image feature space to the label embedding space (F$\rightarrow$E); iii) mapping an unseen image to the label embedding space using the learned mapping function and predicting its label. In this way, ZSL is posed as a {\em missing label problem}. Many existing methods of this type (e.g., \cite{socher2013zero}\cite{al2016recovering}\cite{qiao2016less}) assume a global linear mapping F$\rightarrow$E between the two spaces. \cite{romera2015embarrassingly} present a very simple ZSL approach using this assumption, and extend the approach to a kernel version. However, the global linear mapping assumption can be over-simplified. \cite{wang2016relational} propose to utilize local relational knowledge to synthesize virtual unseen image data so as to simulate the manifold structure of unseen classes, but then back to the global linear assumption to learn the mapping F$\rightarrow$E using both the seen data and synthesised unseen data. We observe that the synthesized manifold structure of unseen classes is not accurate, in addition, back to the global linear mapping assumption further damage the ZSL performance. Hence adaptation should be introduced to adjust the synthesized manifold structure according to the real unseen data.
%simulate the manifold structure of unseen classes

Accordingly, many transductive ZSL approaches are proposed for alleviating the domain adaptation problem\cite{fu2015transductive}. In transductive ZSL, (unlabeled) real unseen data are utilized for refining the trained model, e.g., the label embedding space and mapping function F$\rightarrow$E. \cite{li2015semi} propose a semi-supervised method to learn new label embeddings using prior knowledge of the original ones. In \cite{kodirov2015unsupervised}, a dictionary for the target domain (unseen classes) is learned using regularised sparse coding, and the dictionary learned on the source domain (seen classes) serves as the regularizer. In \cite{zhang2016SPZSL}, a structured prediction approach is proposed. Several clusters on unseen data are generated using K-means, then a bipartite graph matching between these clusters and labels is optimized based on the learned similarity matrix on seen data.
%a bipartite graph matching between seen image clusters and labels is learned, then unseen image data are clustered, and the learned matching is applied to match between the unseen clusters and labels, in this way the unseen images obtain their labels.

Most aforementioned methods aim at learning a potentially complex mapping from F$\rightarrow$E. Under circumstances such as the number of classes is large and there exists polysemy in text labels, such many-to-one ``clean mapping" can be hard to learn. In this paper, we study a novel transductive zero-shot learning method (shown in Figure.\ref{fig:firstfig}), which transfers the manifold structure in the label embedding space to the image feature space (E$\rightarrow$F), and adapts the transferred structure according to the underlying data distribution of both seen and unseen data in the image feature space. As the proposed method associates data to the label, we categorize it as a {\em missing data method} in contrast to the conventional {\em missing label methods}.

%We use the signature to represent the data distribution of each class\cite{romera2015embarrassingly}, which is also referred as prototype\cite{fu2015transductive}.

Our method is based on two assumptions, i) data of each class in the image feature space follow a Gaussian distribution, ii) the local manifold structure of label embeddings are approximate to that of ``the signatures" in the image feature space. In previous works, the signature\cite{romera2015embarrassingly} or prototype\cite{fu2015transductive} is used to denote the authentic distribution of data of each class in the label embedding space. While, in our reverse mapping, we use the ``signature" to
denote the authentic distribution of data of each class in the image feature space. Data distributions are modeled by Gaussians, and ``the signatures" are defined as the model parameters of Gaussians. Our method consists of three main steps:

i) The signature of each seen class is estimated in the image feature space.

ii) The manifold structure is estimated in the labeling embedding space, and is transferred to the image feature space so as to synthesize virtual signatures of the unseen classes in the image feature space.

iii) The virtual signatures are refined, at the same time, each unseen instance is associated to an unseen label (label prediction) by the Expectation-Maximization (EM) algorithm.

Experiments show that the proposed method achieves the state-of-the-art performance on two popular datasets, namely, the Animals with Attributes and the Caltech-UCSD Birds-200-2011. It outperforms the runner-up by nearly \textbf{5\%} and \textbf{10\%} on default and random splits, respectively.

%------------------------------------------------------------------------
\section{The Proposed Method}

$N^s$ seen classes data are denoted as $(\bm{X}^s,\bm{Y}^s)=\{(\bm{x}^s_1,{y}^s_1),...,(\bm{x}^s_{N^s},{y}^s_{N^s})\}$, and $N^u$ unseen classes data are denoted as $(\bm{X}^u,\bm{Y}^u)=\{(\bm{x}^u_1,{y}^u_1),...,(\bm{x}^u_{N^u},{y}^u_{N^u})\}$.
Each datum $\bm{x}^s_{i}$ or $\bm{x}^u_{i}$ $\in \Re^{d \times 1}$ is a $d$-dimensional feature vector in the image feature space.
${y}^s_i$ or ${y}^u_i$ denotes the labels. The label sets of the seen and unseen classes are disjointed, i.e. $ \bm{Y}^s \cap \bm{Y}^u = \emptyset$. The ``auxiliary information" from corpus (e.g. word vectors) or/and annotations (e.g. attributes) are label embeddings denoted as $\bm{E}^s=\{\bm{e}^s_1,...,\bm{e}^s_{K^s}\}$ and $\bm{E}^u=\{\bm{e}^u_1,...,\bm{e}^u_{K^u}\}$ for seen and unseen classes respectively. $\bm{e}^s_{i}$ and $\bm{e}^u_{i}$ $\in \Re^{d' \times 1}$.
Using the seen data pairs $(\bm{x}^s_i,{y}^s_i)$, ZSL aims to predict labels ${y}^u_i$ for each unseen instance $\bm{x}^u_i$ by leveraging the ``auxiliary information" $\bm{E}^s$ and $\bm{E}^u$ for knowledge transfer.

%There exist $N^s$ instances belonging to seen classes data $\bm{X}^s=[\bm{x}^s_1,...,\bm{x}^s_{N^s}]$ and $N^u$ instances belonging to unseen classes data $\bm{X}^u=[\bm{x}^u_1,...,\bm{x}^u_{N^u}]$. Instances $\bm{x}^s_{i} \in \Re^{d \times 1}$ and $\bm{x}^u_{i} \in \Re^{d \times 1}$ are column vectors. Label sets of seen classes  $\bm{Y}^s=[\bm{y}^s_1,...,\bm{y}^s_{K^s}]$ and unseen classes $\bm{Y}^u=[\bm{y}^u_1,...,\bm{y}^u_{K^u}]$ are disjointed, i.e. $ \bm{Y}^s \cap \bm{Y}^u = \emptyset$. Only labels of all seen classes instances are known, i.e. $L(\bm{x}^s_i,\bm{y}^s_j)=1$ if the label of instance $\bm{x}^s_i$ equals $\bm{y}^s_j$, otherwise 0. Zero-shot learning aims to learn the correspondence between each instance and label of unseen classes, i.e. $L(\bm{x}^u_i,\bm{y}^u_j)$. Unless otherwise specified, each instance $\bm{x}^s_{i}$ or $\bm{x}^u_{i}$ refers to its feature vector in the image feature space. Extra knowledge from corpus (e.g. word vectors) or/and manual annotations (e.g. attributes) are utilized as label embeddings $\bm{E}^s=[\bm{e}^s_1,...,\bm{e}^s_{K^s}]$ and $\bm{E}^u=[\bm{e}^u_1,...,\bm{e}^u_{K^u}]$ for seen and unseen classes respectively. $\bm{e}^s_{i} \in \Re^{d' \times 1}$ and $\bm{e}^u_{i} \in \Re^{d' \times 1}$ are column vectors.
%-------------------------------------------------------------------------

\subsection{Estimation of Seen Classes Signatures}
\begin{figure}[htbp]
  \centering
  \includegraphics[width=0.45\textwidth]{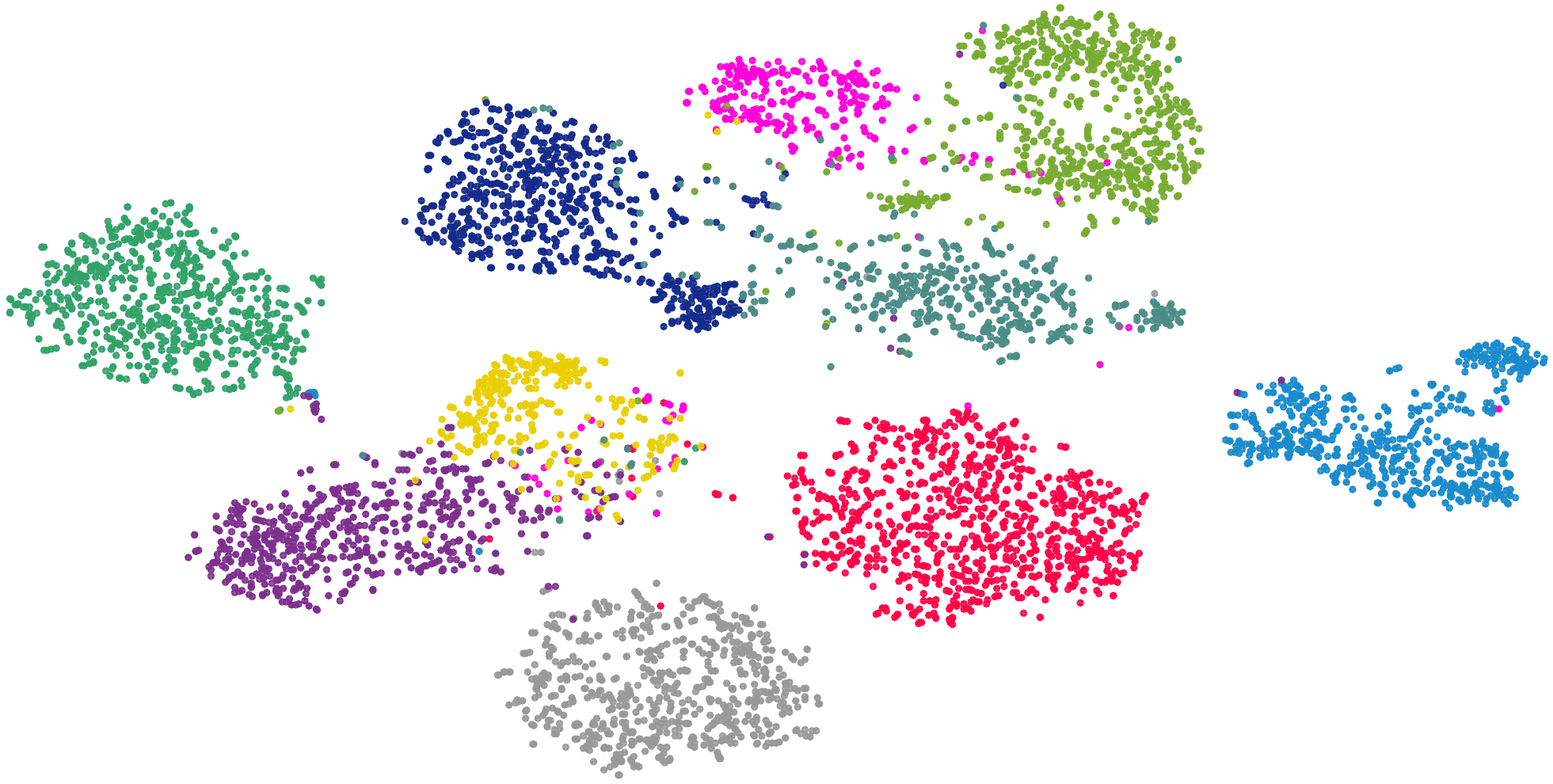}
  \caption{Visualization of the default 10 unseen classes in Animals with Attributes dataset using t-SNE. Instances within each class form a tight cluster.}
  \label{fig:visualization}
\end{figure}

By dimensionality reduction (using t-SNE\cite{maaten2008visualizing}), it is observed that data of each class form a tight cluster (shown in Figure.\ref{fig:visualization}) in the image feature space. Hence, we assume that
\begin{as}
\label{assum1}
Data of each class follow a Gaussian distribution
$\bm{X}\sim {\cal N}(\bm{\mu},\bm{\Sigma})$ in the image feature space.
\end{as}

It is worth noting that in the literature people used Nearest-Neighbor classifiers to assign labels to unseen data, e.g., \cite{palatucci2009zero} \cite{fu2016semi}, the underlying assumption is that the distribution of the data is isotropic Gaussian. Here we estimate the parameters of the Gaussians.

\subsubsection{Estimation of the Signatures}
Similar to \cite{romera2015embarrassingly}, we use ``signature", denoted as $\bm{S}=\{\bm{s}_1,...,\bm{s}_K\}$, to represent the data distribution of each class in the image feature space. The signature is the sufficient statistics of the data, and using it the distribution of the data can be recovered. Here, for a Gaussian model, the signature is $\bm{s}_k =(\bm{\mu}_k ,\bm{\Sigma}_k)$, i.e. the mean and covariance. As the labels of seen classes data are provided, we can estimate signatures of seen classes directly, denoted as $\bm{S}^s$.
%
%Gaussian distribution according to data of each class directly.
%
%establish the GMM on seen classes (sGMM) using labeled data. First, we estimate the separate Gaussian component for each class. Then sGMM can be established by jointing all components.
%
%Here, for Gaussian mixture model, we formulate each signature as $\bm{s}_k =[\pi _k ,\bm{\mu}_k ,\bm{\Sigma}_k ]$, i.e. each Gaussian component. Knowledge transfer from seen classes to unseen classes is performed based on signatures.

\subsection{Synthesis of Virtual Signatures}
One of the key challenges in ZSL is to explore the relationship between the image feature space and the label embedding space. The label embedding is either pre-designed (e.g. by the annotated attribute vectors) or pre-trained on a large corpus (e.g. by word vectors). Although there may not be an accurate global linear mapping from the image feature space to the label embedding space, local manifold structures may be similar of the two.
In this paper we focus on exploiting the local manifold structure rather than the global one. Hence we assume that
\begin{as}
\label{assum2}
The local manifold structure of label embeddings is approximate to that of the signatures in the image feature space and can be transferred for synthesizing the virtual signatures of the unseen classes.
\end{as}
This is formulated as
\begin{equation}
\bm{E}^u =R\left ( \bm{E}^s \right ) \Rightarrow \widehat{\bm{S}^u}=R\left ( \bm{S}^s \right),
\end{equation}
where $\widehat{\bm{S}^u}=\{\widehat{\bm{s}^u_1},...,\widehat{\bm{s}^u_{K^u}}\}$ denotes the synthesized virtual signatures of the unseen classes. There are many choices of the synthesis function $R(\cdot)$ that can approximate the manifold structure of the label embeddings, such as Sparse Coding, K-Nearest Neighbors and so on.

In the literature, many works assume the two spaces observe a global linear transformation so that the structure of the image features can be transferred to the label embeddings via a global linear mapping, e.g., \cite{al2016recovering}\cite{qiao2016less}. We observe that such an assumption is over-simplified. There are works assuming that a global non-linear mapping may exist between the two spaces\cite{romera2015embarrassingly}, e.g., using kernel methods. However, it is prone to get overfitting on the seen data and obtain bad performance on the unseen data. In contrast, our manifold preserving assumption works well empirically in the experiments.
%we observe that it is prone to overfitting (Table.\ref{tab:Art}).

\subsubsection{Synthesis via Sparse Coding}
We choose Sparse Coding\cite{olshausen1997sparse} (inspired by \cite{wang2016relational}) to approximate the manifold structures of the image features and label embeddings. In our implementation, label embeddings of the seen classes serve as the dictionary. Then we compute the sparse linear reconstruction coefficients of the bases for unseen label embeddings.
According to the Sparse Coding theory, we minimize the following loss function to obtain the coefficients $\bm{\alpha}$.
\begin{equation}
\min_{\bm{\alpha} } \left \| \bm{e}^u_k -  \bm{E}^s\bm{\alpha} \right \|^2 + \lambda | \bm{\alpha} |_1,
\end{equation}
where $\bm{\alpha} = {[\alpha_1,...,\alpha_{K^s}]}^T $. This loss function is convex and easy to optimize.

Then, we transfer such local structure from the label embedding space to the image feature space and synthesize the virtual signature of each unseen class using the same set of coefficients, i.e. $  \widehat{\bm{s}^u_k} = \bm{S}^s \bm{\alpha}$, where the components in $\bm{E}^s$ and $\bm{S}^s$ correspond to each other. This transferring is valid because that the distribution of an unseen class in the image space is assumed to be a Gaussian and the components either in $\bm{E}^s$ or $\bm{S}^s$ are assumed to be independent.

After synthesizing all unseen signatures (say $K^u$ of them), the distribution of the unseen instances $\{\bm{x}^u_n\}$ in the image feature space is a Gaussian Mixture Model (GMM),
\begin{equation}
p\left( {\bm{x}^u_n} \right) = \sum\limits_{k = 1}^{K^u} {{\pi_k}{\cal N}\left( {\bm{x}^u_n|{\bm{\mu}_k},{\bm{\Sigma}_k}} \right)}
\label{equ:GMM}
\end{equation}
$\pi _k$ denotes the $k$th mixing coefficient and its initial value is assumed to be $1/K^u$. The initial value of $(\bm{\mu}_k,\bm{\Sigma}_k)=\widehat{\bm{s}^u_k}$. $\bm{x}^u_n$ denotes the $n$th image in $\bm{X}^u$.

The synthesized virtual signatures approximate the distribution of the unseen data in the image feature space. However, they may not be accurate. Next, we optimize/refine the signatures, at the same time, associate each unseen image to an unseen label. This is the reason we pose our ZSL as a missing data problem.

\subsection{Solving the Missing Data Problem}
We impute unseen image labels and update the GMM parameters using the Expectation-Maximization (EM) algorithm.

The objective function is defined as the log of the likelihood function,
\begin{equation}
\ln{p\left( {{\bm{X}^u}|{\bm{\pi}, \bm{\mu}},\bm{\Sigma} } \right)} = \sum_{n=1}^{N^u}{\ln{}} \sum\limits_{k = 1}^{K^u} {{\pi _k}{\cal N}\left( {{\bm{x}^u_n}|{\bm{\mu}_k},{\bm{\Sigma} _k}} \right)}
\end{equation}

In the Expectation step, the conditional probability of the latent variable $y^u_n=k$ given $\bm{x}^u_n$ under the current parameter is
\begin{equation}
p(y^u_n=k | \bm{x}^u_n) = \frac{{\pi _k}{\cal N}\left( {{\bm{x}^u_n}|{\bm{\mu}_k},{\bm{\Sigma} _k}} \right)}{\sum_{j=1}^{K^u}{{\pi _j}{\cal N}\left( {{\bm{x}^u_n}|{\bm{\mu}_j},{\bm{\Sigma} _j}} \right)}}.
\end{equation}
This is the posterior probability of an unseen image $\bm{x}^u_n$ belonging to label $k$.

In the Maximization step, the model updates the parameters using the posterior probability.
\begin{equation}
\bm{\mu}_k^{new} = \frac{1}{N^u} \sum_{n=1}^{N^u}{p(y^u_n=k | \bm{x}^u_n) \bm{x}^u_n}
\end{equation}
\begin{equation}
\bm{\Sigma}_k^{new} = \frac{1}{N^u} \sum_{n=1}^{N^u}{p(y^u_n=k | \bm{x}^u_n) (\bm{x}^u_n-\bm{\mu}_k^{new})^T(\bm{x}^u_n-\bm{\mu}_k^{new})}
\end{equation}
\begin{equation}
\pi^{new}_k = \frac{N^u_k}{N^u}
\end{equation}
where
\begin{equation}
N^u_k = \sum_{n=1}^{N^u}{p(y^u_n=k | \bm{x}^u_n)}
\end{equation}
$K^u$ and $N^u$ denote the number of all unseen classes and instances respectively. We iterate the E-step and M-step until convergence. After the convergence, the parameters of the data distribution are refined and the unseen instances are assigned with labels.

%Correspondence $C^u$ among unseen instances and labels is assigned as the probability of each instance to each signature. The probability of an instance belong to unseen classes is given by the likelihood of the Gaussian Mixture Model. Here we use the same notations as that in [].
%With the established model, we assign the label of unseen instance $X^u_i$ as $Y^u_k$, s.t. $k = \arg \max_k C^u_{ik}$.

\subsubsection{Regularization}
\label{sec:analysisOptimization}
During the EM process when estimating the GMM, each covariance matrix $\bm{\Sigma}_k$ should be nonsingular, i.e. invertible. For a reliable computation, empirically, the number of data in each class $N_k$ should be greater than the square of feature dimension, i.e. $\forall_k, N_k \geq \lambda d^2, s.t. \; \lambda \geq 1 $. $\lambda$ is a coefficient. However, this may not be satisfied in some situations when feature dimension is high but only a small number of data are provided per class.

We employ two tricks to solve this problem, namely, dimensionality reduction and regularization of $\bm{\Sigma}_k$. For dimensionality reduction, we choose to use linear dimension reduction methods, e.g. principal components analysis (PCA), to reduce the image feature representation to $d$ dimensional, which is much smaller than the original one.

If we only choose to stabilize the computation by reducing the image feature dimension, the label prediction accuracy will degrade quickly. Hence,  we also resort to another solution, i.e., regularizing $\bm{\Sigma}_k$. Here, we present two regularization methods of $\bm{\Sigma}_k$, namely, diagonal $\bm{\Sigma}_k$, $s.t. \; N_k \geq \lambda d$ and unit $\bm{\Sigma}_k$, $s.t. \; N_k \geq 1$. Diagonal $\bm{\Sigma}_k$ means that $\bm{\Sigma}_k$ is assumed to be a diagonal matrix. Unit $\bm{\Sigma}_k$ means that $\bm{\Sigma}_k$ is an identity matrix. These two regularization methods simplify $\bm{\Sigma}_k$ in an increasing order. We choose to use a simpler one if the number of the data is smaller.

%\begin{equation}
%\begin{array}{c}
%\forall_k,\quad N_k \geq \lambda d^2 \\
%s.t. \quad \lambda \geq 1
%\end{array}
%\end{equation}

%-------------------------------------------------------------------------
\section{Experiments}
\subsection{Datasets \& Settings}
In this section, we evaluate the proposed method by conducting experiments on two popular datasets, i.e., the Animals with Attributes (AwA) \cite{lampert2009learning} and the Caltech-UCSD Birds-200-2011 (CUB) \cite{WahCUB_200_2011}.

AwA\footnote{http://attributes.kyb.tuebingen.mpg.de/} contains 50 classes and 85 manual attributes (both binary and continuous). The average number of the images of each class is 610, and the minimum number is 92. Ten classes serve as the unseen classes and the remaining forty are utilized as the seen classes. \cite{lampert2014attribute} provided a fixed default split, which is used as the default split in many works.

CUB\footnote{http://www.vision.caltech.edu/visipedia/CUB-200-2011.html} is a fine-grained image dataset which contains 200 species of birds annotated with 312 binary attributes.  The mean and minimum numbers of bird images of each class are 60 and 41 respectively. Commonly, 50 species are chosen as the unseen classes, and the rest are the seen classes. The fixed default split used in this paper follows that in \cite{wang2016relational}.

For AwA, we use i) 4096-dimensional VGG features (VGG-fc7) provided along with the dataset, ii) 1024-dimensional GoogLeNet features, iii) 1000-dimensional ResNet features. For CUB, we use iv) 1024-dimensional GoogLeNet features, v) 1000-dimensional VGG features (VGG-fc8) and vi) 2048-dimensional ResNet features extracted on the Pooling-5 layer. ii, iii, iv, v) are provided by \cite{wang2016relational}. The label embeddings (attributes and word vectors) used in this paper are the same as \cite{wang2016relational}.

Most previous works presented their experimental results using a fixed default split or a few random splits of the seen/unseen classes  on different datasets. We argue that the evaluation based on the fixed default split or only a few random splits may not be comprehensive/stable enough, especially on small-scale datasets. For a fair comparison, we evaluate our method on both ``many random splits" and the fixed default split. ``Many random splits" means that we conduct all experiments with 300 random splits.

\subsection{Analysis of Data Distribution}
First, we examine if \textbf{Assumption \ref{assum1}} is a reasonable assumption, i.e. the data of each class approximately subject to a Gaussian distribution in the image feature space. The idea is to show that under this assumption the upper bound of the proposed ZSL performance exceeds that of the state-of-the-art methods by a considerable margin.

To obtain the upper bound performance of the proposed method under \textbf{Assumption \ref{assum1}}, we conduct a upper-bound experiment, in which the labels of all data (both seen and unseen) are given. Hence, we can estimate the Gaussian distribution for each class according to the data labels. Then the label of each datum is predicted as the one with the maximum likelihood of the Gaussians/classes. The mean classification accuracy consequently can be computed.

%\subsubsection{Image Feature Space}
Table.\ref{tab:assum1} shows the upper-bound classification performances of the proposed method based on \textbf{Assumption \ref{assum1}} in different image feature spaces. All-50 means that we estimate Gaussian distributions on all 50 classes of AwA and report the overall classification accuracy. Unseen-10 means we estimate Gaussians on 10 randomly selected classes as unseen classes and the classification accuracy is the average over 300 such random trials. All-200 and Unseen-50 have the similar meanings for CUB dataset.

For all classes of AwA, modeling data with Gaussian achieves 84.55\% classification accuracy in VGG-fc7 feature space. For all classes of CUB, the classification accuracy is 73.81\% in GoogLeNet+ResNet feature space.

The experimental upper bound performance under \textbf{Assumption \ref{assum1}} on AwA Unseen-10 and CUB Unseen-50 are 92.10\% and 85.03\% using VGG-fc7 and GoogLeNet + ResNet features respectively. According to Table.\ref{comp}, the proposed upper-bound performance is much larger than the corresponding state-of-the-art performance \--- 68.05\% (RKT) and 61.69\% (RKT) on AwA and CUB respectively. Therefore, the Gaussian assumption of the distribution of data is reasonably good currently when comparing the proposed method with the other state-of-the-arts.

It is worth noting that it is reasonable for CUB to have a lower upper-bound than that of AwA, as CUB is a fine-grained bird species dataset, hence the classification is harder.

% \usepackage{multirow}
\begin{table}[ht]
\centering

\begin{tabular}{c|c|c|c}
\hline
\multicolumn{1}{c|}{} & \multicolumn{1}{c|}{Image Feature} & \multicolumn{1}{c|}{Setting} & Acc. \% \\ \hline
\multirow{4}{*}{AwA}  & \multirow{2}{*}{VGG-fc7}       & All-50                       & 84.55       \\ \cline{3-4}
                      &                              & Unseen-10                    & 92.10       \\ \cline{2-4}
                      & GoogLeNet                    & All-50                       & 81.44       \\ \cline{2-4}
                      & ResNet                          & All-50                       & 73.51       \\ \hline
\multirow{4}{*}{CUB}  & \multirow{2}{*} {GoogLeNet + ResNet}           & All-200                       & 73.81       \\ \cline{3-4}
                      &                              & Unseen-50                    & 85.03       \\ \cline{2-4}
                      & {GoogLeNet}   & All-200                       & 67.48       \\ \cline{2-4}
                      & GoogLeNet + VGG-fc8           & All-200                       & 60.43       \\ \hline
\end{tabular}
\caption{Analysis of data distribution assumption. All-50 and Unseen-10 means that we estimate the GMM on all 50 classes and random 10 unseen classes of AwA respectively. All-200 and Unseen-50 have the similar meanings for CUB. The high classification accuracies explain that \textbf{Assumption \ref{assum1}} is effective in different feature spaces and datasets.}
\label{tab:assum1}
\end{table}

\subsection{Effectiveness of Virtual Signatures}
\label{sec:recon}
To justify \textbf{Assumption \ref{assum2}}, we evaluate the classification performance using synthesized virtual signatures directly. This strategy can be viewed as inductive ZSL. We run 300 random trials on AwA and CUB respectively. Features extracted from VGG-fc7 (4096-dim) for AwA and GoogLeNet+ResNet (3072-dim) for CUB are utilized. We use the same label embeddings as those in \cite{wang2016relational}. According to our analysis in Sec.\ref{sec:analysisOptimization}, the image feature dimension is reduced to 80-dim on AwA. Because the minimum number of images of each class is 92. We also reduce the feature dimension of CUB data to 400-dim for speeding up the computation. Three types of label embedding are tested, namely, attributes(A), word vectors(W) and attributes with word vectors(A+W). Results using different settings are shown in Table.\ref{tab:SigEM}.

As shown in Table.\ref{tab:SigEM}, the classification accuracies using synthesized signatures without EM step are 72.11\% on AwA and 59.94\% on CUB (using A+W label embeddings), which is comparable to the sate-of-the-art (see Table.\ref{comp} and Table.\ref{tab:Art}). These results show that the synthesized signatures are reasonably good and so is \textbf{Assumption \ref{assum2}}.

We find that the performance using word vectors (60.99\%) as label embedding is better than that using attributes (58.73\%) on AwA. However, this phenomenon reverses on CUB (i.e. 47.31\% using word vectors and 56.21\% using attributes). A possible reason is that the general training corpus for the word vector model is not specific to fine-grained bird species. So word vectors of fine-grained species names do not work well as those of the general animal names.

\subsection{Evaluation of the EM Optimization} \label{sec:EM}
Here, we evaluate the gain brought by the EM optimization (shown in Table.\ref{tab:SigEM}). All data (features, label embeddings, random splits) are consistent with those in the previous subsection. GMM with diagonal $\bm{\Sigma}_k$ (GMM-EM-Diagonal) and unit $\bm{\Sigma}_k$ (GMM-EM-Unit) are tested. For AwA, GMM-EM-Unit brings about 17\%  improvement of classification accuracy using the three label embeddings on average. Using GMM-EM-Diagonal increases nearly 1\% classification accuracy over the GMM-EM-Unit. For CUB, nearly 6\% improvement is brought by using GMM-EM-Unit. The experiment using GMM-EM-Diagonal on CUB is not reported due to the lack of training data (about 60 data in each class, which is explained in Sec.\ref{sec:analysisOptimization}). These results show that the EM optimization improves classification performances in different settings.

We also implement a baseline algorithm to show the effectiveness of using synthesized signatures as the initialization of the EM optimization as shown in Table.\ref{tab:SigEM}. In Baseline-Random-Init.-EM, we randomly pick a set of unseen datapoints to initialize the mean of the GMM components, then execute the EM optimization. The resulted classification accuracies are 9.46\% on AwA and 2.00\% on CUB respectively, which are at chance level.

%In Baseline-Groundtruth, we assume that groundtruth labels of these clusters are given, then we can accurately estimate the GMM parameters. Performances of our method (87.38\% on AwA and 63.37\% on CUB) are close to those of this baseline algorithm (87.76\% on AwA and 60.49\% on CUB). The reason why our method has a slightly better result on CUB than the baseline algorithm may be the limitation of modeling the fine-grained CUB dataset using the GMM.

% \usepackage{multirow}
\begin{table*}[]
\centering
\begin{tabular}{c|c|c|c|c|c}
\hline
\multirow{2}{*}{}    & \multirow{2}{*}{\begin{tabular}[c]{@{}c@{}}Label\\ Embedding\end{tabular}} & \multirow{2}{*}{\begin{tabular}[c]{@{}c@{}}Acc. \% of\\ Syn.-Sig\end{tabular}} & \multirow{2}{*}{\begin{tabular}[c]{@{}c@{}}Acc. \% of\\ GMM-EM-Unit\end{tabular}} & \multirow{2}{*}{\begin{tabular}[c]{@{}c@{}}Acc. \% of\\ GMM-EM-Diagonal\end{tabular}} & \multirow{2}{*}{\begin{tabular}[c]{@{}c@{}}Baseline\\ Random-Init.-EM \end{tabular}} \\
                     &                                                                            &                                                                                &                                                                                   &                                                                                       &                                                                                                                                                                  \\ \hline
\multirow{3}{*}{AwA} & A                                                                          & 58.73                                                                          & 82.44                                                                             & 83.39                                                                                 & \multirow{3}{*}{9.46}                                                                                                                       \\ \cline{2-5}
                     & W                                                                          & 60.99                                                                          & 75.31                                                                             & 76.31                                                                                 &                                                                                                                                                                   \\ \cline{2-5}
                     & A+W                                                                        & 72.11                                                                          & 86.39                                                                             & 87.38                                                                                 &                                                                                                                                                                   \\ \hline
\multirow{3}{*}{CUB} & A                                                                          & 56.21                                                                          & 61.27                                                                             & \multirow{3}{*}{-}                                                                    & \multirow{3}{*}{2.00}                                                                                                                       \\ \cline{2-4}
                     & W                                                                          & 47.31                                                                          & 55.62                                                                             &                                                                                       &                                                                                                                                                                   \\ \cline{2-4}
                     & A+W                                                                        & 59.94                                                                          & 63.37                                                                             &                                                                                       &                                                                                                                                                                  \\ \hline
\end{tabular}
\caption{Evaluate the synthesized virtual signatures with and without the EM optimization algorithm under the 300 random split setting. Syn.-Sig. denotes classification directly using the synthesized virtual signatures. GMM-EM-Diagonal and GMM-EM-Unit are two regularization methods that use diagonal $\bf{\Sigma}_k$ and unit $\bf{\Sigma}_k$ in the EM algorithm to estimate the GMM. Using GMM-EM with unit $\bm{\Sigma}_k$ brings about 17\% and  6\%  improvement on AwA and CUB respectively. On AwA, using GMM-EM with diagonal $\bm{\Sigma}_k$ increases nearly 1\% classification accuracy over the one using the unit $\bm{\Sigma}_k$. The last column shows that if we initialize the GMM component using random datapoints, the classification accuracy is at chance level.}
\label{tab:SigEM}
\end{table*}

\subsection{Comparison to the State-of-the-Art}

First, we compare our method to two popular methods, namely ESZSL \cite{romera2015embarrassingly} and RKT \cite{wang2016relational}, using provided codes. We repeat these experiments using the same setting (including image features, label embeddings, the default split and 300 random splits ) as the aforementioned in Sec.\ref{sec:recon}. Although we have to reduce image feature dimensions in our method, we use the original image features for other methods.

From Table.\ref{comp}, it can be seen that on AwA the average classification accuracy of our method is \textbf{87.38\%}, which outperforms that of the runner-up (RKT) 68.05\% by {19.33\%} on the random splits. On CUB, the performance of our method is \textbf{63.37\%}, which also exceeds that of the runner-up (RKT) 61.69\% by {1.68\%} on the random splits. This superiority is also observed on the default split setting on two datasets. We use the same set of model parameters for both the default and random split settings, rather than using different parameters on different settings. The inductive version of our method (Ours\_I) achieves comparable results on the two split settings on two datasets.

\begin{figure}[t]
  \centering
  \includegraphics[width=0.42\textwidth]{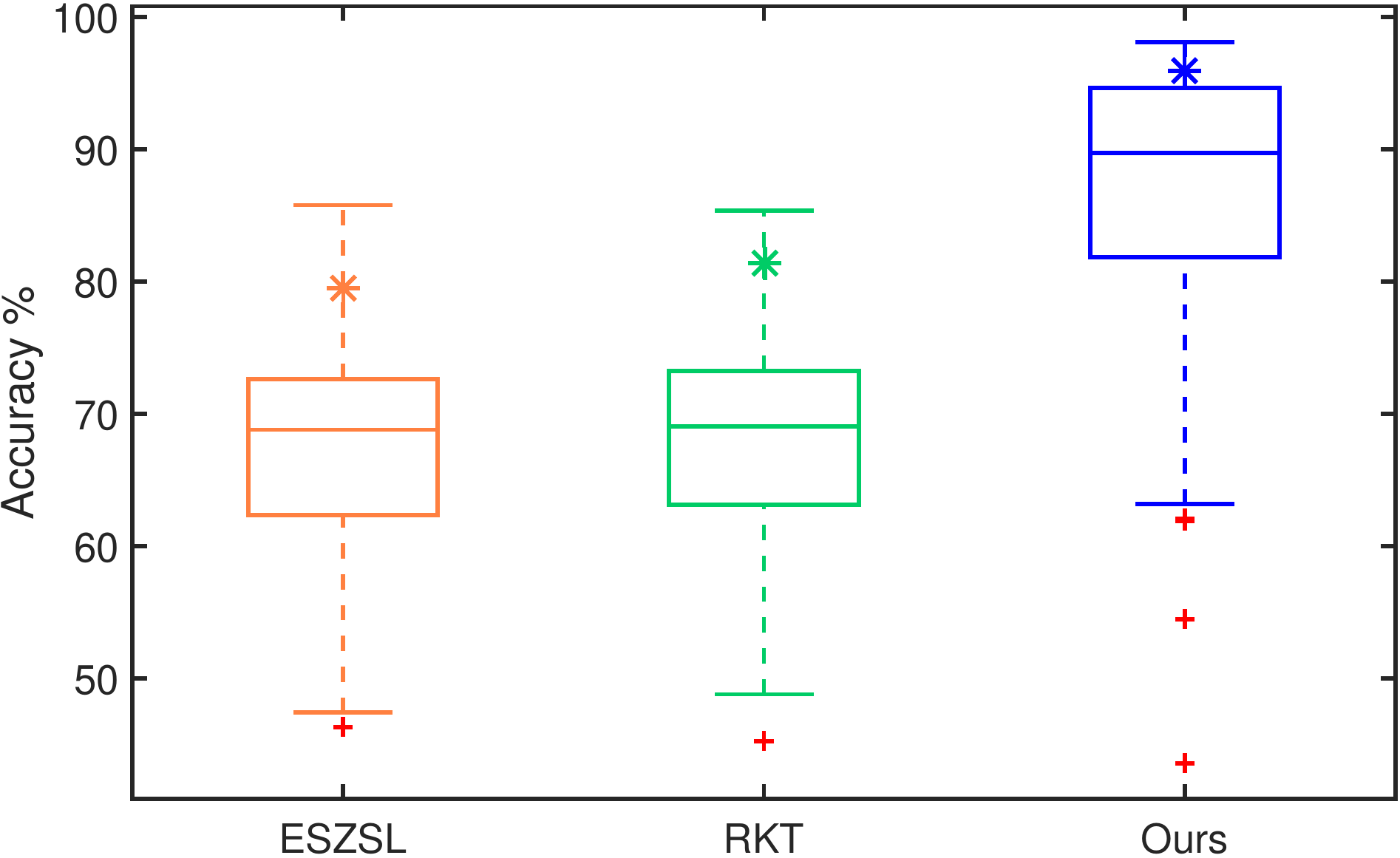}
  \caption{Box plots of results of different methods on AwA. The three box plots illustrate the classification accuracy statistics of the ESZSL, RKT and our method on 300 random splits. On each box, the central mark is the median, the edges of the box are the 25th and 75th percentiles. It is shown that the variance of random split results is large. The stars represent results of the three methods on the default split.}
  \label{fig:var}
\end{figure}

From Figure.\ref{fig:var} we find that the variance of the random split classification accuracies is large for all the three methods on AwA. By contrast, the classification accuracies of the default split (marked as stars in the figure) are all in good positions in the performance bars. This supports our argument that the experiments on large number of random splits are necessary for reliable results and comparison.

% \usepackage{multirow}
\begin{table}[]
\centering
\caption{Comparison to popular methods using the same setting. Ours\_I is our method using the synthesized virtual signatures directly for classification without the EM optimization.}
% Our method achieves highest performances on both AwA and CUB.
%, including image features, label embeddings (A+W) and 300 random splits
\label{comp}
\begin{tabular}{c|c|c|c|c}
\hline
                     &   \multirow{2}{*}{Method}     &    \multirow{2}{*}{Image Feature}    &    \multicolumn{2}{c}{Accuracy \%}          \\
                     &   &                                                & Default &Random \\ \hline
\multirow{4}{*}{\begin{turn}{90}AwA\end{turn}} & ESZSL   & \multirow{4}{*}{VGG-fc7}                        &  79.53   & 67.75       \\ \cline{2-2} \cline{4-5}
                     & RKT     &                                                                           &  81.41  & 68.05       \\ \cline{2-2} \cline{4-5}
                     & Ours\_I &                                                                           &  82.07  & 72.11       \\ \cline{2-2} \cline{4-5}
                     & Ours    &                                                                            & \textbf{95.99}  & \textbf{87.38}       \\ \hline
\multirow{4}{*}{\begin{turn}{90}CUB\end{turn}} & ESZSL   & \multirow{4}{*}{\begin{tabular}[c]{@{}c@{}}GoogLeNet\\ + ResNet\end{tabular}}                                                         &    51.90       & 57.62       \\ \cline{2-2} \cline{4-5}
                     & RKT     &                                                                    &55.59  & 61.69       \\ \cline{2-2} \cline{4-5}
                     & Ours\_I &                                                                        & 57.31  & 59.94       \\ \cline{2-2} \cline{4-5}
                     & Ours    &                                                                  & \textbf{60.24}  & \textbf{63.37}       \\ \hline
\end{tabular}
\end{table}

We also compare with the results reported in recent papers, namely DAP/IAP \cite{lampert2014attribute}, ESZSL \cite{romera2015embarrassingly}, SJE \cite{akata2015evaluation}, SC\_struct \cite{changpinyo2016synthesized}, SS-Voc \cite{fu2016semi}, JLSE \cite{zhang2016zero}, Mul-Cue \cite{akata2016multi}, TMV-HLP \cite{fu2014transductive}, RKT \cite{wang2016relational}, SP-ZSR \cite{zhang2016SPZSL} and LatEm \cite{xian2016latent}. From Table.\ref{tab:Art}, it can be seen that our method achieves the best performance on the both datasets.

From Table.\ref{tab:Art}, it can be seen that on AwA our method achieves the best accuracy on the default split, i.e. \textbf{95.99\%}, which is 3.91\% improvement compared to the runner-up method, i.e. 92.08\% of SP-ZSR. There are few works, namely LatEm, SC\_struct and DAP/IAP, evaluated on random splits, but only on a few random trials. We evaluate our method on 300 random trials and achieve \textbf{87.38\%} classification accuracy on AwA. Our result is almost 11.28\% higher than that of the runner-up, LatEm.

From Table.\ref{tab:Art}, it can be seen that the average performance on CUB is not as good as that on AwA. This is also observed in the previous experiments. Our method achieves \textbf{60.24\%} classification accuracy on the default split, which outperforms the runner-up (SP-ZSR) by 4.90\%. Notice that the classification accuracy of 56.5\% achieved by Mul-Cue requires manual annotation for the bird part locations of the test images. So, it is not fair to compare with this result directly. Our method receives \textbf{63.37\%} mean accuracy on the 300 random splits. This result is 8.67\% higher than the runner-up (SC\_struct). Overall, our method achieves nearly 5\% and 10\% improvement on the default and random splits respectively compared to reported results on the both datasets.
% As fine-grained datasets of birds, it is even hard for ordinary people (without training) to classify different bird species.

% \usepackage{multirow}
\begin{table}[t]
\centering
\begin{tabular}{c|c|c|c|c}
\bottomrule
                         & Methods                     & Split                                                          & AwA                & CUB            \\ \toprule
\multirow{8}{*}{\begin{turn}{90}Default\end{turn}}
                         & DAP/IAP                    &                                                                & 41.4/42.2         & -              \\ \cline{2-2} \cline{4-5}
                         & ESZSL                      &           \multirow{8}{*}{Default}                          & 49.30              & -              \\ \cline{2-2} \cline{4-5}
                         & SJE                 &                                                                & 66.7             & 50.1          \\ \cline{2-2} \cline{4-5}
                         & SC\_struct                &                                            & 72.9               & -          \\ \cline{2-2} \cline{4-5}
                         & SS-Voc                      &                                                                & 78.3               & -              \\ \cline{2-2} \cline{4-5}
                         & JLSE                       &                                                        & 80.46              & 42.11          \\ \cline{2-2} \cline{4-5}
                         & \multirow{2}{*}{Mul-Cue} &                                                                & \multirow{2}{*}{-} & 56.5*          \\ \cline{5-5}
                         &                             &                                                                &                    & 43.3           \\ \cline{2-2} \cline{4-5}
                         & TMV-HLP                    &                                                                & 80.5               & 47.9           \\ \cline{2-2} \cline{4-5}
                         & RKT                        &                                                & 82.43              & 46.24          \\ \cline{2-2} \cline{4-5}
                         & SP-ZSR                &                                                                & 92.08              & 55.34          \\ \cline{2-2} \cline{4-5}
                         & Ours                        &                                                                & \textbf{95.99}     & \textbf{60.24} \\ \midrule

\multirow{6}{*}{\begin{turn}{90}Random\end{turn}}
                         & DAP/IAP                    & 5 trials                                                       & 37.1/34.1         & -              \\ \cline{2-5}
                         & LatEm                      & \begin{tabular}[c]{@{}l@{}}1 default\\ + 4 trials\end{tabular} & 76.1               & 47.4           \\ \cline{2-5}
                         %& JLSE\cite{zhang2016zero}                        & 3 trials                                                       & 80.46              & 42.11          \\ \cline{2-5}
                         & SC\_struct                & 4 trials                                                         & -               & 54.7           \\ \cline{2-5}
                         & Ours                        & 300 trials                                                     & \textbf{87.38}     & \textbf{63.37} \\ \toprule
\end{tabular}
\caption{Comparison to the state-of-the-art. * means extra information is used. On AwA, our method outperforms the runner-ups by 3.91\% and 11.28\% on the default and random splits respectively. On CUB, our method outperforms the runner-ups by 4.90\% and 8.67\% on the default and random splits respectively.}
\label{tab:Art}
\end{table}

%--------------------------------------------------------------------------
\section{Conclusion}
In this paper, we propose a transductive zero-shot learning method based on the estimation of data distribution by posing ZSL as a missing data problem. Different from others, we focus on exploiting the local manifold structure in two spaces rather than the global mapping. Testing data are classified in the image feature space based on the estimated data distribution. Experiments show that the proposed method outperforms the state-of-the-art methods on two popular datasets.
%Our method has better simplicity and interpretability than others.

\bibliographystyle{named}
\bibliography{ijcai17}

\end{document}